\documentclass{article}
    \usepackage{arxiv}
    \usepackage{multirow}
    \usepackage{moreverb,url}
    \usepackage{amsmath,amssymb}
    \usepackage{multicol}

\usepackage{stmaryrd}
\usepackage{booktabs}
\usepackage{graphicx}

    \newcommand{\botrule}{\bottomrule}
    \newcommand{\tbl}{\caption}
    \newcommand{\colrule}{\midrule}
    \renewenvironment{multline}{\begin{equation}\begin{gathered}}{\end{gathered}\end{equation}}

\usepackage[noibid,authordate,backend=biber]{biblatex-chicago}
\begin{document}

\title{Anomaly Detection And Classification In Time Series With Kervolutional Neural Networks}

    \author{%
    	Oliver Ammann\\
        ETH Zürich,\\
        Zurich, Switzerland\\
        \And 
        Gabriel Michau\\
        ETH Zürich,\\
        Zurich, Switzerland\\
        \And 
        Olga Fink\\
        ETH Zürich,\\
        Zurich, Switzerland}
        \subtitle{ESREL 2020 - PSAM 15}
        \date{14th of May 2020}
        \header{O. Ammanm et al. - ESREL 2020 - PSAM 15}
        \maketitle

\begin{abstract} 
Recently, with the development of deep learning, end-to-end neural network architectures have been increasingly applied to condition monitoring signals. They have demonstrated superior performance for fault detection and classification, in particular using convolutional neural networks. Even more recently, an extension of the concept of convolution to the concept of kervolution has been proposed with some promising results in image classification tasks. 
In this paper, we explore the potential of kervolutional neural networks applied to time series data. We demonstrate that using a mixture of convolutional and kervolutional layers improves the model performance. The mixed model is first applied to a classification task in time series, as a benchmark dataset. Subsequently, the proposed mixed architecture is used to detect anomalies in time series data recorded by accelerometers on helicopters. We propose a residual-based anomaly detection approach using a temporal auto-encoder. We demonstrate that mixing kervolutional with convolutional layers in the encoder is more sensitive to variations in the input data and is able to detect anomalous time series in a better way.
\end{abstract}


\section{Introduction}

Efficient operation of industrial systems requires safe and  highly available assets. Avoiding failures is, therefore, of utmost importance. To prevent failures due to excessive wear, the operators typically perform the maintenance based on manufacturers' recommendations, and have to inspect  the critical parts on a regular basis. This is an expensive and time-consuming process. On the contrary, performing the maintenance based on the condition of the system could benefit the operation by delaying unnecessary maintenance and preventing evolving faults. Since monitoring sensors are increasingly becoming  cheaper, they have been increasingly widely used in mechanical systems. Condition monitoring sensors combined with the connectivity and transmission of the measured signals have enabled real-time monitoring of a growing number of systems. This provides an opportunity to gain information about the operation of their systems and to plan future operation and maintenance accordingly. The analysis of the data enables not only to detect anomalies but also to isolate faults. Yet, current approaches often rely on handcrafted features which are often not robust to changes in the operating conditions and the evolution of new faults. Additionally, feature engineering often encounters scalability problems, particularly when the amount of collected data increases. 

As a consequence, there is a strong need to automate anomaly detection (AD) in condition monitoring data. Automatic AD algorithms exist in several fields such as in the medical domain, computer vision, Internet of Things architectures and in complex industrial systems \parencite{Chalapathy2019}. However, few methodologies have been developed for safety-critical systems with high-frequency multivariate time series data. 

For such systems, reliable AD algorithms require efficient and effective handling of multivariate time series. Previous approaches have been mainly based on extracting features on   which the AD algorithms are trained. This requires the design of relevant statistical or spectral features to be used in subsequent tasks, such as hidden Markov models~\parencite{li2017multivariate} or clustering algorithms~\parencite{aghabozorgi2015time}. Designing the right features for each problem and adjusting the models accordingly needs a lot of expertise from the engineers. An end-to-end anomaly detector that can be directly trained on raw condition monitoring data  would, therefore, be beneficial and would significantly improve the scalability . 

Over the last years, deep learning has become popular for its capacity to train end-to-end models from massive datasets with little feature engineering required. However, handling multivariate, long and high-frequency time series with deep learning approaches is still a challenging task\cite{kanarachos2017detecting, wang2020deep}. After the huge success of convolutional neural networks (CNN's) in the domain of image processing, the CNN's have also proven to be successful in anomaly detection \parencite{he2019temporal, Zhang2019, Zhang2017, Ince2016}.

Recently, an extension of convolution has been introduced called kervolution \parencite{Wang2019}. The research study has shown that the use of kervolution benefits the model capacity for image classification tasks. Kervolution replaces the convolution operation by a non-linear operation, increases the learning capacity of the model and leads to models with faster convergence.

In this paper, we demonstrate that the use of kervolutional layers also benefits time series related tasks. We first demonstrate improvement in a time series classification task. Then we also use kervolutional layers on two anomaly detection tasks for time series. 

The remainder of the paper is organized as follows: Section~\ref{ch::method} presents the concept of kervolution, the kernels explored in the present work and the design of a kervolutional layer. Section~\ref{ch::classification} demonstrates the benefits in using kervolution within a classification task on the Human Activity Recognition dataset (HAR). Section~\ref{ch::AD} presents the results obtained on two case studies for Anomaly Detection. 

\section{Methodology}
\label{ch::method}
    \subsection{From convolution to kervolution}
In deep learning, convolutional layers perform the convolution operation between the input of the layer, denoted by $X\in\mathbb{R}^{L,W}$ and a matrix $V$, of pre-defined size $M\times N$, whose elements $V_{m,n}\in\mathbb{R}$ are to be learned.
Usually, we have $M\leq L$ and $N\leq W$, and $V$ is denoted as ``filter'' by analogy with the filtering operation in signal processing. 

The convolution comprises the following operation:
\begin{multline}
\label{eq:convol}
\forall (i,j)\in (\llbracket M,L\rrbracket\times\llbracket N,W\rrbracket), \\ f(i,j) = \sum_{m=1}^M \sum_{n=1}^{N} X_{i-m,j-n}\cdot V_{m,n}.
\end{multline}
Eq.~\ref{eq:convol} can be rewritten as the inner product of $X$ and $\hat{V}$ where $\hat{V}$ is $V$ inverted : $\hat{V}_{m,n} = V_{M-m,N-n}$. Then we have
\begin{multline}
\label{eq:convol-inner}
\forall (i,j)\in (\llbracket M,L\rrbracket\times\llbracket N,W\rrbracket),\\ f(i,j) = \langle X_{((i-M)\rightarrow i),((j-N)\rightarrow j)} , \hat{V} \rangle.
\end{multline}

In traditional convolutional layers, these operations are generalised to a collection of matrix $V$, with pre-defined sizes. If all matrices in the collection do not have the same size, the layer is denoted as \emph{inception} layer \parencite{szegedy2015going}. Without loss of generality, the concept can be simplified to the one-dimensional convolution ($N=1$) or extended to higher dimensions.

Inspired by the kernel-trick developed for Support Vector Machine \parencite{boser1992training,scholkopf2002learning}, \cite{Wang2019} propose to extend the concept of convolution using kernels such as:
\begin{equation}
\label{eq:convol-kervol}
\begin{split}
\forall (i,j) & \in (\llbracket M,L\rrbracket\times\llbracket N,W\rrbracket),\\
f(i,j) & = \langle \phi(X_{((i-M)\rightarrow i),((j-N)\rightarrow j)}) , \phi(\hat{V}) \rangle\\
 & = \kappa(X_{((i-M)\rightarrow i),((j-N)\rightarrow j)},\hat{V}),
\end{split}
\end{equation}

where $\phi$ is a non-linear mapping function and $\kappa$ is a kernel function. \cite{Wang2019} demonstrate that it maintains two important properties of the convolution: (1) weight sharing (contrary to a dense layer where each element in $X$ is assigned a different set of weights), and (2) equivariance to translations in $X$. It brings also increased model capacity with new non-linear features. The convolutional layer can be considered as a special case of a kervolutional layer with a linear kernel.

    \subsection{Selection of the Kernels}
In addition to the hyper-parameters already existing for the convolutional layer, such as the size of the filter $V$ ($M$ and $N$), the number of filters used, the possible use of bias or regularisation of the weights, the kervolutional layer also requires the choice of a suitable kernel. The original contribution \parencite{Wang2019} does not provide any recommendation on the the choice of the kernel \textit{a priori}, that is, without using a validation dataset. 

Note that to increase the performance of the Convolutional Neural Networks, a non-linear activation function is applied to the results of the convolution operations. Since kervolution already performs non-linear operations within the kernel, \cite{Wang2019} claim that the use of activation functions is not required anymore. In that sense, the choice of the activation function has been replaced by the choice of a suitable kernel, which keeps the number of hyper-parameters identical. 

In this project we focus on the polynomial kernel proposed in the work of \cite{Wang2019}. 
The polynomial kernel leads to the easiest implementation, the fastest training with performances that we could not outperform with other kernels in our preliminary research. These findings are consistent with those of \parencite{Wang2019}. 

    \subsection{The Polynomial Kernel}

The polynomial kernel performs the following operation:
\begin{equation}
\label{eq:ploykern}
\kappa \left( x,v \right) = \left( x^Tv + c_p \right)^{d_p}.
\end{equation}
where $c_p$ is a bias and $d_p$ is the degree of the kernel. 
 
Our preliminary results demonstrated that for $d_p>1$ and  using several consecutive layers of kervolution, the output of the neurons has a tendency to diverge. Therefore, in this project, we propose to systematically use a batch-normalisation layer after a kervolutional layer. As a side remark, we also implemented a small modification of the kernel, to fix this behavior in some cases: 
\begin{equation}
\kappa_p \left( x,v \right) = \left( \dfrac{x^Tv}{K_p} + c_p \right)^{d_p}
\label{eq::ploykernnorm}
\end{equation}
where $K_p$ acts as a normalization constant, eg. $K_p = card(x)$.

    \subsection{Metrics}
In this paper we evaluate the performance of our models with the following metrics: the Accuracy, the Precision and the Recall, also denoted as True Positive Rate (TPR). We denote respectively by $TP$, and $FP$ the number of true and false positive samples, by $TN$ and $FN$ the number of true and false negative samples. Then we have:

\begin{equation}
\begin{aligned}
Accuracy &= \dfrac{TP+TN}{TP+TN+FP+FN} \\
Precision &= \dfrac{TP}{TP+FP}\\
TPR &= \dfrac{TP}{TP+FN}
\end{aligned}
\end{equation}

\section{Kervolution for Time-Series Classification}
\label{ch::classification}

\subsection{Benchmark Case study: The Human Activity Recognition Dataset}

To evaluate the benefit of using kervolution instead of convolution for time series data, we first evaluate the approach on a benchmark dataset, the human activity recognition (HAR) dataset \parencite{Anguita2013}. The HAR dataset consists of measured acceleration and velocity of a human, recorded with Galaxy S2 smartphones, in the three dimensions. This results in a six-dimensional dataset. The original task proposed with the release of this dataset is the classification and recognition of the activity of the owner of the smartphone \parencite{Anguita2013}. This task is relevant from an industrial perspective since accelerometers are used in many industrial systems~\parencite{ompusunggu2019long,zhang2019deep}. They are relatively cheap to build and can be retro-fitted. They can be used to detect, recognise and classify operating conditions, system wear or faults~\parencite{wang2020missing}. 

The HAR dataset consists in 10299 samples, each 2.56 seconds long and sampled at 50 Hz, that is, 128 measurements per sample. The authors of the original study who released the dataset \parencite{Anguita2013} already split the dataset for training and testing with a ratio of around 70\% and 30\%. They also applied a low-pass Butterworth filter with cutoff frequency at 20Hz for noise reduction. We applied in addition a standardisation of each variable at the dataset level, whose statistics have been computed on the training set.
The data is labelled with the six classes detailed in Table~\ref{tab::HAR_class}. The dataset is almost balanced.

\begin{table}
\centering
\tbl{Number of samples per class in the HAR dataset}
{\tabcolsep14pt
\begin{tabular}{ll}
\toprule
Categories & \# of samples\\ \midrule
Walking & 1722\\
Walking upstairs & 1544\\
Walking downstairs & 1406\\
Sitting & 1777\\
Standing & 1906\\
Laying & 1944\\ 
\bottomrule
\end{tabular}}
\label{tab::HAR_class}
\end{table}

\subsection{Protocol}
The best classification results achieved on the HAR dataset based on raw signals (without any feature extraction) are reported in \parencite{Ronao2016}. In that study, a three-layer deep convolutional neural network was applied. In our experimental setup, we first reproduced the exact same architecture with convolutional layers as reported in \parencite{Ronao2016}. In the second step, we found a simpler network architecture that could achieve a similar performance as the architecture reported in \parencite{Ronao2016}. In the third step, we evaluated the benefits of replacing convolution with kervolution. To enable a fair comparison, we use the same architecture as in the pure CNN experiment but replace the convolutional layers with kervolution within this simpler network. 

The network reported in \parencite{Ronao2016} has the following settings:
\begin{itemize}
    \item 3 convolutional layers with 94, 192 and 192 filters of size 9$\times$1, and ReLu as activation function, each followed by a max pooling layer of size 3$\times$1,
    \item a dropout layer with a dropout rate of 0.8,
    \item a 1000-neurons dense layer,
    \item a 6-neurons dense layer with softmax activation function.
\end{itemize}
The network is trained with Stochastic gradient descent (SGD), with a learning rate of 0.02, a momentum of 0.5 and a decay of 5e-05. The training uses a max number of epochs of 5000 and early stopping with a patience of 100 epochs and retains the  weights of the best performance.

Our simpler network (with a similar performance as the original architecture) has 16, 8 and 4 filters instead, no dropout layer and no intermediate dense layer. 

All our networks are trained with the Adam optimizer, a learning rate of $10^{-3}$ and a batch size of 128. Training lasts for 100 epochs and early stopping with a patience of 20 epochs.

To quantify the benefits of using kervolution instead of convolution, we replace the convolutional layers by kervolutional layers in the simpler network. We test all possibilities for every layer between a convolutional layer with a ReLu activation function, and a kervolutional layer of degree between $2$ and $4$. We keep the best model after 5-fold cross-validation (80\%/20\%) on the training set and we report the results for the test dataset.

\subsection{Results}

In Table \ref{tab::class_res}, we report the results for a) the original network architecture, b) the simplified network architecture, and c) the network architecture as in b) but with a combination of kervolutional and convolutional layers. For all cases, the results after 5-fold cross validation are reported. We can observe that the best model using kervolution outperforms the CNN versions. The best performing architecture comprises a kervolutional layer with a polynomial kernel of second degree followed by a convolutional layer and a kervolutional layer with a polynomial kernel of third degree. We refer to it as the ``mixed network'' since it combines convolutional and kervolutional layers.

\begin{table}
\tbl{Results of the different classification models on the HAR dataset}
{\tabcolsep6pt
\begin{tabular}{@{}ll@{}}\toprule 
Models & Accuracy \\ \colrule

a) Pure CNN (architecture as in \cite{Ronao2016}) & 89.94\%  \\
b) Pure CNN (simplified) & 89.27\%  \\
c) Mixed network (same architecture as in b) & 91.73\%  \\ \botrule

\end{tabular}}
\label{tab::class_res}
\end{table}

\begin{figure}
\centering
\caption{Confusion matrix of the ``mixed network'': The two classes sitting and standing are the hardest to separate.}
\includegraphics[width=6cm]{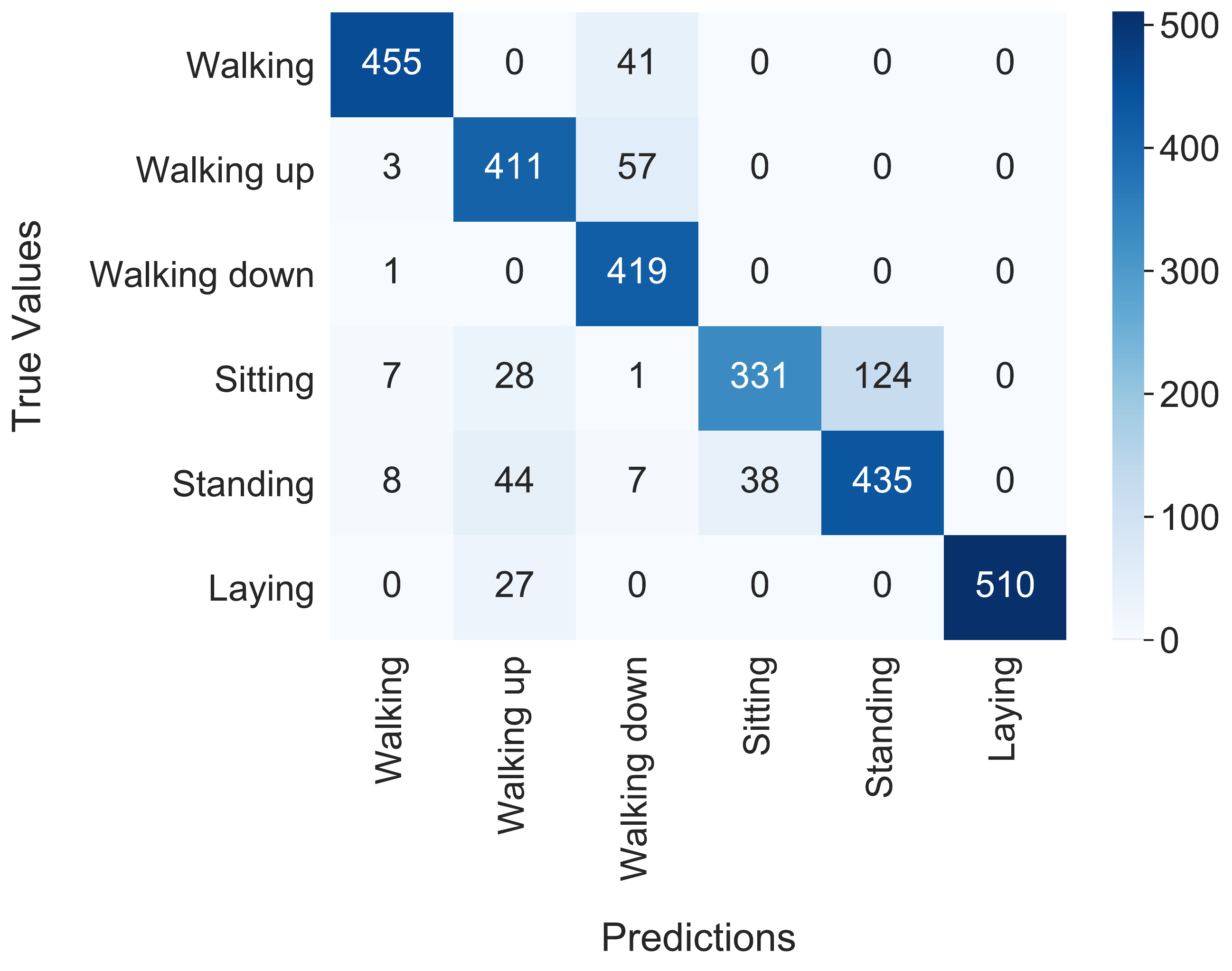}
\label{fig::conf_matrix}
\end{figure}

For the mixed network, the confusion matrix is displayed in Figure \ref{fig::conf_matrix}. The model achieves very good detection performances on almost all classes except for the sitting and standing classes. To the best of our knowledge, all other classifiers in the literature meet the same difficulty due to the high similarity between these two classes.

\section{Kervolution for Anomaly Detection}
\label{ch::AD}
\subsection{Anomaly Detection Algorithm}
To identify abnormal data we use a residual-based AD algorithm. This algorithm first reconstructs the signal with an autoencoder. Then, the residual is computed as the average over the time series of the absolute differences between the reconstructed and the original signal. 
Using normal data that has not been used for training, a threshold is set at the 99th percentile of its residuals. 
To test a sample, it is fed through the auto-encoder and its residual is computed. If its residual exceeds the threshold, it is detected as abnormal.

Similarly to the classification experiments performed in Section \ref{ch::classification}, we first evaluate an autoencoder with convolutional layers, and then use the same architecture but replace these layers with kervolutional layers to evaluate the performance of kervolutional neural networks. 
Our autoencoders are designed as follows: each con- or kervolutional layer is followed by a max pooling layer. Please note that in the decoder, only \textbf{deconvolutional} layers followed by unpooling layers are used since "de-kervolutional" layers have not yet been proposed. The corresponding operation to the deconvolution for kervolution is still an open research question.

We train the autoencoder on a healthy dataset to reconstruct the input as accurately as possible. The loss used for its training is the root mean square error between the original signal and its reconstruction.

\subsection{Anomaly detection based on Human Activity Recognition}

As the first experiment for anomaly detection, we propose to formulate the HAR classification task as anomaly detection. Since the activities of ``sitting'' and ``standing'' cannot be distinguished by a supervised classifier, we combine them to a single class. In Figure \ref{fig::conf_matrix} we see that these two classes even when combined have the highest confusion rate, we take them as the main class and will consider all the other classes as anomalies of different types. 
To compare the performance of the AD algorithm, a custom metric called balanced accuracy (BA) is used. The balanced accuracy is a weighted average of the true positive rates (TPR) of each class and the false positive rate (FPR) of the classes that have been considered as normal. The equation for the BA is provided in the following:

\begin{equation}
\begin{aligned}
& BA = \left( \sum_{j \in J_2} j + \left( 100-fpr \right) \cdot |J_2| \right) \cdot \dfrac{1}{2\cdot  |J_2|}, \\
& where \; J_2 = \{tpr_1, tpr_2, tpr_3, tpr_6\}
\end{aligned}
\end{equation}

$Tpr_i$ is the true positive rate of the i-th fault type and $|J_2|$ corresponds to the number of tested fault types. In our case $|J_2|=4$. 

\subsubsection{Protocol}

Similarly to the classification task, we start with a convolutional neural network, and will explore the impact of replacing its layers with kervolutional layers. We built a simple convolutional autoencoder that minimises the residuals on a validation healthy set.

Consequently, the autoencoder that is used for this problem has only one layer in the encoder and in the decoder. This model which is referred as CNN in table \ref{tab::res_AD_HAR} has a convolutional layer with 4 kernels of size 4x6, a stride of 2 and a tanh activation function, followed by a max pooling layer with a kernel size of 2x1 and a stride of 2 in the encoder. The decoder uses an upsampling layer with a kernel size of 2x1 followed by a deconvolutional layer with 4 kernels of size of 8x6 and a stride of 2. 
The best model uses a kervolutional layer with a polynomial kernel of third degree followed by a batch normalization layer instead of the convolutional layer in the encoder. 

\subsubsection{Results}

The results in Table \ref{tab::res_AD_HAR} indicate that both models are in fact perfect at detecting the other classes. However, the KCNN model has a lower FPR. It might come from the higher learning capacity of kervolution which allowed to learn the main class more efficiently.

\begin{table}
\centering
\tbl{Results of the autoencoder, using either a convolutional layer in the encoder or a kervolutional layer. Numbers with a * are false positive rates since it is the normal class (NC).}
{\tabcolsep14pt
\begin{tabular}{@{}lll@{}}\toprule 
                    & CNN           & KCNN    \\ \colrule
                    & TPR (\%)      & TPR (\%)      \\ \colrule
Class 1             & 100           & 100           \\
Class 2             & 100           & 100           \\
Class 3             & 100           & 100           \\
Class 6             & 100           & 100           \\
NC                  & 2.37*         & 1.53*          \\ \colrule
Balanced Accuracy   & 98.81         & 99.24         \\ \botrule
\end{tabular}}
\label{tab::res_AD_HAR}
\end{table}

\subsection{Anomaly detection on Helicopter Flight Test}

To test the kervolution on a real application, a univariate dataset collected from helicopters has been used. The dataset consists of measured vibrations at different spots on the helicopters. Neither the orientation, nor the aircraft of origin, nor the location of the measurements is known. Additionally, the whole dataset has been scaled by an unknown factor. The data has been recorded at a frequency of 1024 Hz. Each sample has a length of one minute which results in 61440 measurements per sample. The data set consists of 2271 samples. The predefined training set includes 1677 samples, where all of the samples are considered as normal. The test set includes 594 samples, half of the samples are normal and half of them are abnormal. To the best of our knowledge, there are currently no published results on this dataset in the literature.

We performed a minimal data pre-processing on this dataset. We split the samples into sub-sequences of length 512, corresponding to half a second. This pre-processing step has two advantages: first, it reduces the input size and the training time; second, it allows for a finer analysis of the time series. By analyzing at a finer time scale, a transition from healthy to unhealthy can be captured by successive sub-sequences. This leads to 201\,240 sequences in the training set. For the testing, if at least one of the sub-sequences is detected as anomalous, the whole sample is labeled as abnormal. 

\subsubsection{Protocol}
Similarly to the previous experiment, we designed the convolutional autoencoder such that it achieves the lowest residuals on the training set, using grid search. Consequently, the autoencoder used here consists of three layers in the encoder and in the decoder. 
The convolutional layers in the encoder have 32, 64, 128 kernels and kernel sizes of 16, 8, 4 from the input to the latent space. The strides are 4, 2, 2 accordingly. We use a tanh function as activation. Each convolutional layer is followed by a max pooling layer with a kernel size of 2 and a stride of 2. The latent space size is 160. The decoder is composed of three blocks, each with one upsampling layer and one deconvolutional layer. All upsampling layers have a kernel size of 2 and the deconvolutional layers have 64, 32, 1 kernels. The sizes of the kernels are 4, 8, 16 and the strides are 2, 2, 4 accordingly.

To compare the fully convolutional autoencoder to the kervolutional autoencoders, the convolutional layers in the encoder have been replaced by kervolutional layers. In the following, we present the results for the model where all convolutional layers in the encoder have been replaced by kervolutional layers. The following models are used: 
\begin{itemize}
    \item fully convolutional autoencoder (CNN)
    \item kervolutional neural network with a polynomial kernel of third degree followed by batch normalization (KCNN-d3)
     \item kervolutional neural network with a polynomial kernel of second degree and batch normalisation (KCNN-d2)
     \item kervolutional neural network with a polynomial kernel of third degree and the normalization proposed in equation \ref{eq::ploykernnorm} (KCNN-Kp\&BN)
    \item kervolutional neural network with a polynomial kernel of third degree as in  KCNN-Kp\&BN but without batch normalization layers after the kervolutional layers (KCNN-Kp).
\end{itemize}


\subsubsection{Results}

The results of the different models show that the kervolutional autoencoder with a polynomial kernel of third degree provides the best results compared to all the other models (Table \ref{tab::1D_res}). The models using kervolutional layers have higher precision, probably due to the higher learning capacity of kervolutions compared to the convolution-based models. The results indicate that the use of both the batch normalization and the normalised kernel performs worse than using only one of the two regularization techniques. However, if the normalization techniques are used alone, they achieve a higher TPR than the baseline CNN model.

\begin{table}
\centering
\tbl{Results of the different models on the helicopter flight test data}
{\tabcolsep6pt
\begin{tabular}{@{}llll@{}}\toprule 
Models & TPR & Precision & Accuracy\\ \colrule

Model CNN & 60\% & 100\% & 79.9\% \\

Model KCNN-d3 & 68.7\% & 100\% & 84.3\% \\

Model KCNN-d2 & 67.0\% & 100\% & 83.5\% \\

Model KCNN-Kp\&BN & 60.3\% & 99.4\% & 80.0\% \\

Model KCNN-Kp & 66.0\% & 100\% & 83.0\% \\ \botrule
\end{tabular}}
\label{tab::1D_res}
\end{table}

\section{Discussion}

In this paper, we demonstrated on three experiments that  kervolutional neural networks or mixed architectures provide better classification and anomaly detection performance compared to pure convolutional neural networks.

First, in the classification task, mixing kervolution and convolutional layers leads to better performance, compared to the same model that only uses convolutional layers. In this specific task, a mix of convolutional and kervolutional layers with polynomial kernels of different degrees has proven to perform the best. 

Second, in two different anomaly detection tasks, the models including kervolutional layers have performed better than the neural network with the same architectures using only convolutional layers. Particularly in the helicopter case study, the improvement of the model using kervolutional layers is significant. In both AD experiments, the improvements rely on a decrease in the false positive rate, demonstrating a higher learning capacity of kervolutions.

The results of the AD task with the helicopter data have also shown that the use of normalization in the polynomial kernel leads to acceptable results, yet it could not outperform the model using batch normalization instead. On the other hand, a combination of both the batch normalization layers and the normalized kernel harms the performance of the models. It is important to mention, however, that the normalization of the kernel function has not been extensively tuned or analyzed.

The main focus of this paper has been on the comparison of convolution and kervolution within similar architectures. We did not aim to reach the best possible score for the tasks but rather ensured a fixed and reproducible framework to compare the kervolution with the convolution. As a consequence, a deeper analysis of the optimisation tasks could provide more suitable and more performant architectures to the different tasks. However, we demonstrated that after optimising the architecture in a way that CNN provides reasonable results, using non-linear kernels instead of the classic convolution improves the performance. Therefore, kervolution should be considered as an option when handling time series.

\section{Conclusion}

In this paper, we evaluated the benefits of using kervolution over convolution for time series classification and anomaly detection. We demonstrated that kervolution efficiently extends the concept of convolution since the convolution can be seen as a kervolution with a polynomial kernel of degree one. We demonstrated that mixing kernels of varied degrees outperforms pure convolutional architectures both in classification tasks and in AD tasks. Our results also illustrated the higher learning capability of such models. In anomaly detection task, the main class is more robust and the false positive rate is lower. In fact, false alarms are a major obstacle to the implementation of the end-to-end machine learning approaches in industry. Since the model interpretation is difficult, the false alarms are hard to diagnose. If they are too frequent, the cost associated with the analysis of false alarms can rapidly exceed the benefits of using an end-to-end learning approach. 

Future works will focus on the development of de-kervolution. An extensive evaluation of other kernels in the scope of kervolution also requires further research. This includes the question of polynomial kernel normalisation to avoid diverging outputs. Additionally, future work could focus on the development of recommendations for selecting the most adapted kernels and their combinations depending on the application. Last, online implementation of the methodology, to pinpoint the time of the transition between healthy and abnormal behaviour would be a valuable addition.

\section*{Acknowledgement}
This research was funded by the Swiss National Science Foundation (SNSF) Grant no. PP00P2 176878.

\printbibliography

\end{document}